\documentclass[10pt,twocolumn,letterpaper]{article}

\usepackage{cvpr}      % To produce the REVIEW version
\usepackage{graphicx}
\usepackage{amsmath}
\usepackage{amssymb}
\usepackage{adjustbox}
\usepackage{booktabs}
\usepackage{multirow}
\usepackage{color}
\usepackage{verbatim}

\usepackage[pagebackref,breaklinks,colorlinks]{hyperref}

\usepackage[capitalize]{cleveref}
\crefname{section}{Sec.}{Secs.}
\Crefname{section}{Section}{Sections}
\Crefname{table}{Table}{Tables}
\crefname{table}{Tab.}{Tabs.}

\def\cvprPaperID{*****} % *** Enter the CVPR Paper ID here
\def\confName{ECCV TiE workshop}
\def\confYear{2022}

\begin{document}

\title{Character decomposition to resolve class imbalance problem in Hangul OCR}

\author{Geonuk Kim$^{1}$\qquad Jaemin Son$^{2}$\qquad Kanghyu Lee$^{2}$\qquad Jaesik Min$^{1}$\\
$^{1}$42dot.ai \qquad $^{2}$Hyundai Motor Group\\ 
{\tt\small \{geonuk.kim, jaesik.min\}@42dot.ai\qquad \tt\small \{woalsdnd, kanghyulee\}@hyundai.com}
}

\maketitle

%%%%%%%%% ABSTRACT
\begin{abstract}
% As a phonogram, Hangul can represent 11,172 different characters only with 52 graphemes, by describing each character with a combination of the graphemes.
We present a novel approach to OCR(Optical Character Recognition) of Korean character, Hangul. As a phonogram, Hangul can represent 11,172 different characters with only 52 graphemes, by describing each character with a combination of the graphemes. 
As the total number of the characters could overwhelm the capacity of a neural network, the existing OCR encoding methods pre-define a smaller set of characters that are frequently used. This design choice naturally compromises the performance on long-tailed characters in the distribution. In this work, we demonstrate that grapheme encoding is not only efficient but also performant for Hangul OCR. Benchmark tests show that our approach resolves two main problems of Hangul OCR: class imbalance and target class selection. 
\end{abstract}

%%%%%%%%% BODY TEXT
\section{Introduction}
\label{sec:intro}
Optical Character Recognition (OCR) technology aims to recognize texts in the form of pixels in an image, which has long been studied in the field of computer vision. 
% Recent advances in deep learning have shown dramatic performance improvements thanks to ever-growing model sizes, acceleration of hardware, and massive amount of publicly available datasets. 
Most research efforts\cite{hu2020gtc,shi2016end,he2016reading,graves2006connectionist,cnn_base,yu2020towards} have been dedicated to the recognition of English characters which consists of 26 classes of the alphabet and few efforts\cite{chinese1,chinese2,indic} address other languages. 
% Techniques that applied for English characters may not directly apply to Korean Hangul where the total number of recognition targets incurs severe class-imbalance in appearances.
However, the techniques applied to English character recognition cannot be directly applied to Hangul where the total number of recognition targets induces a severe class-imbalance.
%  the total number of recognition targets induces a severe class-imbalance에 대한 부가설명 없어도 되는지...

%Previous architectures for the text recognition task can be broken down into three parts - a Convolutional Neural Network (CNN) for feature extraction of input images, a Recurrent Neural Network (RNN) for modeling the sequence of characters, and a Fully-Connected (FC) layer for the final classification with CTC loss\cite{graves2006connectionist}.

\begin{figure}
\begin{center}
\includegraphics[width=0.8\linewidth,height=3.0cm]{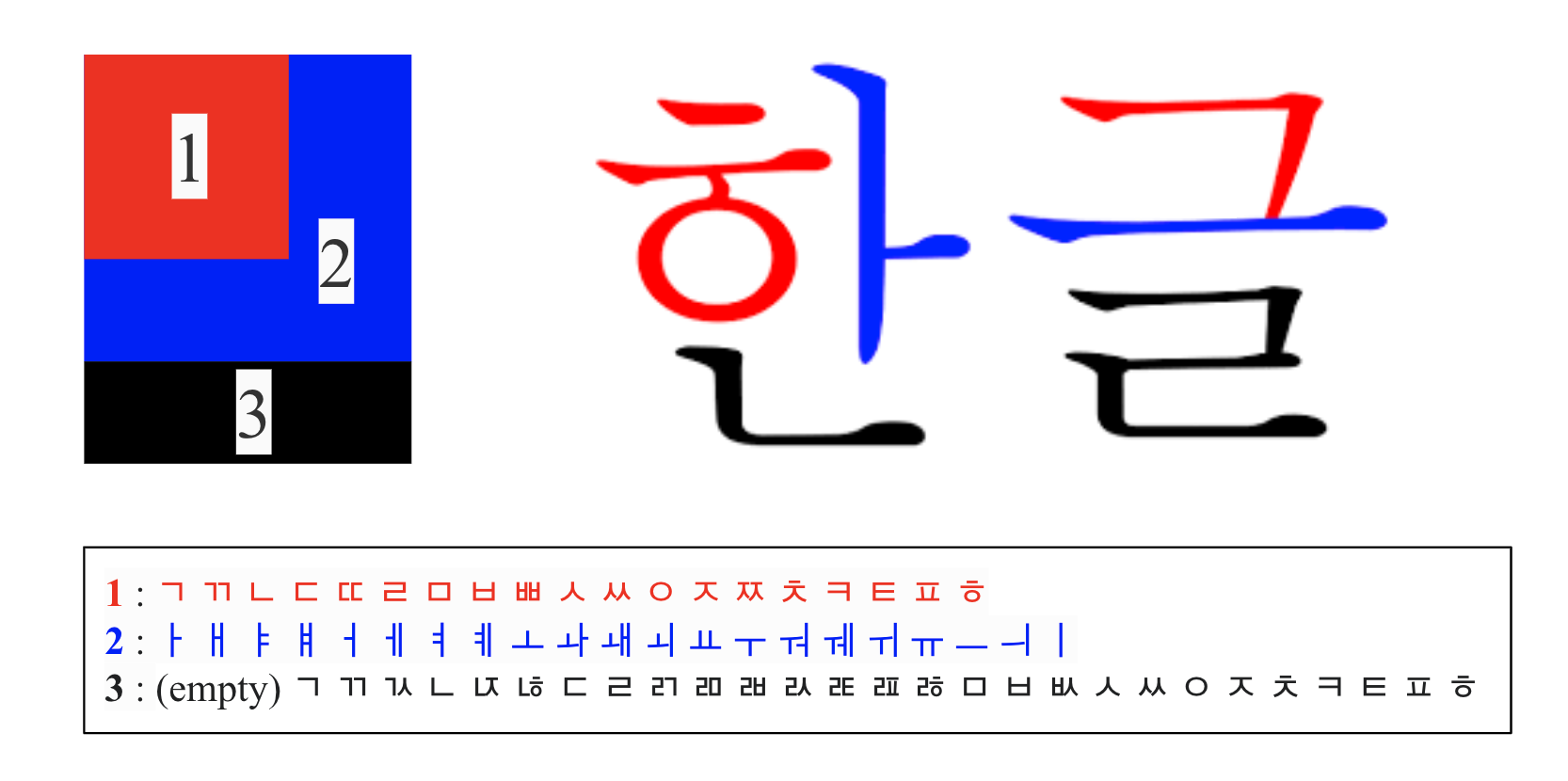}
\end{center}
\caption{Innate structure of the Hangul, Korean character. A character is composed of 2 or 3 graphemes either [\textit{first consonant + vowel}] or [\textit{first consonant + vowel + last consonant}]. The total number of legitimate characters amounts to 11,172.} \label{fig:hangul_structure}
\end{figure}

Unlike English, each Hangul character is described as a combination of 2 or 3 graphemes as shown in Figure \ref{fig:hangul_structure}: either [\textit{first consonant + vowel}] or [\textit{first consonant + vowel + last consonant}]. The total number of characters amount to 11,172, which may easily exhaust softmax layers in representational capability. 
Hence, the current Hangul OCR systems utilize only a subset of characters, maintaining about a thousand most frequently used ones. 
Therefore, the encoding choice naturally excludes characters that would appear in reality. Moreover, the dataset is severely disproportionate in terms of the appearance of characters, which aggravates the class-imbalance issue. Previous approaches attempt to generate virtual images to provide evenly-distributed-characters\cite{kim2015handwritten,kang2018study,park2019study}, however, they cause extra training time for learning characters that may not be critical to target domains.
% (e.g. driving environment, documents, and business name).
 \begin{figure*}[h]
\begin{center}
\includegraphics[width=1.0\linewidth,height=3.5cm]{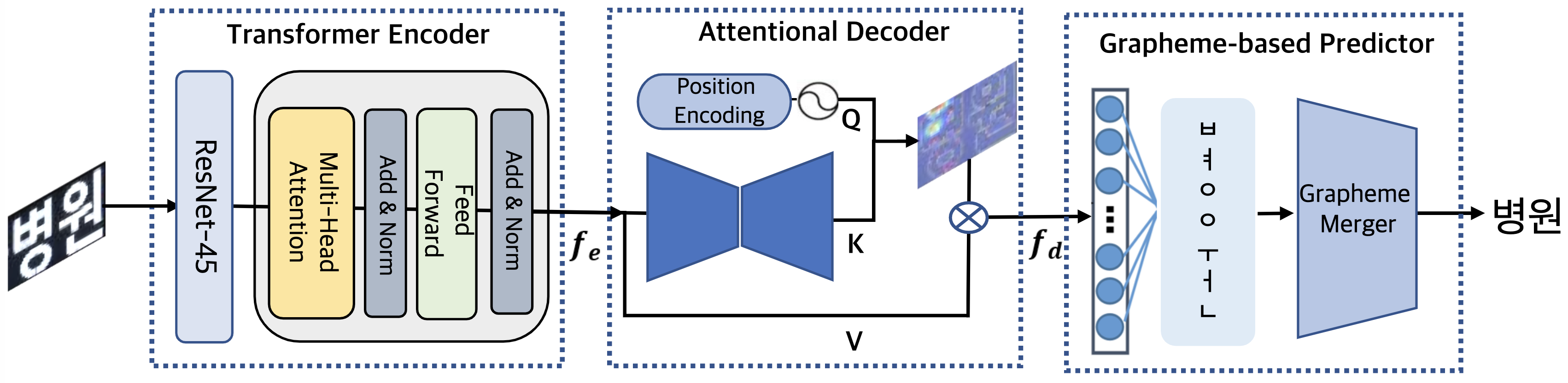}
\end{center}
\caption{An overview of the proposed method. Transformer-based backbone encodes non-local feature representation and decoder attends to feature corresponding to each grapheme. Finally, grapheme-level classification covered and it is merged in a form of character.  }\label{fig:architecture}
\end{figure*}

To address this issue, we propose to decompose the character into graphemes.
Since consonants and vowels are at each position with fixed rules in Hangul ([\textit{first consonant, vowel}] or [\textit{first consonant, vowel, last consonant}]), changing the recognition target from characters to graphemes can replace the problem of classifying 11,172 classes to 52 classes. 
% ?
This decomposition has two main advantages: first, the class-imbalance problem is significantly alleviated as we let the model learn to classify the graphemes in the majority class and minority class of characters into the same grapheme class. 
Second, unlearned characters can also be recognized through the combination of previously learned consonants and vowels, enabling character recognition for the entire characters of Hangul. We construct a benchmark for Korean text recognition task and experimentally verified the effectiveness of the proposed method.  
%이 부분 abstract에 요약해서 쓰면 어떨까?

\section{Related Work}
Scene text recognition task, as a subtask of the entire OCR system, aims to recognize texts in a cropped image that primarily contains text objects. The techniques are widely used in various real world applications ranging from license plate recognition, cursive recognition, to credit card number recognition~\cite{memon2020handwritten}.

Canonical methods modeled the recognition task as a sequence prediction - visual features are extracted from a CNN backbone such as resnets~\cite{he2016deep} and fed into RNN layers such as LSTM~\cite{hochreiter1997long} and GRU~\cite{cho2014properties} to be trained with connectionist temporal classification (CTC) loss~\cite{graves2006connectionist}. 
Recent approaches~\cite{fang2021read, huang2022swintextspotter} make use of more robust backbone such as vision transformer (ViT)~\cite{dosovitskiy2020image} that models interactions between image patches, inspired by attention mechanisms in Transformer~\cite{vaswani2017attention} developed in natural language processing. 
% Transformer backbone eliminates the needs of RNN layers that the network can be trained with cross entropy loss instead of computation-heavy CTC loss.  

% As alphabet has only 26 entities, previous researches handling English characters do not seriously suffer from class-imbalance. However,
% Hangul has 11,172 characters, which may be a burden for the classification. 
% Choosing which characters to train from 11,172 characters in Hangul has been handled in different angles in Korean text recognition.
The selection of which of the 11,172 characters of Hangul to train has been handled from different angles in Korean text recognition.
% The choice of characters is not a trivial decision in Korean text recognition. 
Previous studies bifurcate into 1) selecting only frequently used characters~\cite{park2019study,kim2015handwritten} or 2) including all of the 11,172 characters~\cite{kang2018study}. Both of the methodologies have obvious limitations: in the former approaches, characters outside the chosen set can never be recognized. 
The latter requires much more data containing a wider range of characters, in order to classify characters that rarely appear in real-world situations.
Theoretically, the optimal sets of characters can be searched for all specific domains, yet, it is resource-intensive and still unclear whether the trained models are transferable between different domains.

\section{Method}
For scene text recognition in Hangul, the input is an image and the output is the text in Hangul. 
As shown in Figure \ref{fig:architecture}, we employ an encoder-decoder structure which is widely used in previous studies\cite{cnn_base,transformer_base,yu2020towards,fang2021read}. In encoder, CNN layer $\phi$  and transformer-unit $\psi$  are employed to learn the context between graphemes that define a character. For the input image $\textbf{x}$, we have: 
\begin{equation}
\textit{$f$}_\textit{e} = \textit{$\psi$ (\textit{$\phi$ \textit{(\textbf{x})}})}\quad{ \in R^{\frac{H}{4}\times\frac{W}{4}\times{N}}}
\end{equation}
where \textit{N} is the feature dimension and \textit{H},\textit{W} are the size of   \textit{\textbf{x}}.

Then the decoder learns to attend feature corresponding to each grapheme in the given rich feature map $f_e$.
As shown in Figure \ref{fig:hangul_structure}, each grapheme exists in certain geometric prior. Thus, we express the positional relationship among each set of [\textit{first-middle-last}] graphemes on the feature map by using positional encoding. With the position encodings of the character orders $Q\quad{ \in R^{L \times{C}}}$ and length of the total sequence \textit{L},
\begin{equation}
\textit{$f$}_\textit{d} = \textit{softmax}(\frac{QK^T}{\sqrt{N}})V\quad{ \in R^{L\times N}}
\end{equation}
where $K\quad{ \in R^{\frac{H}{4}\times\frac{W}{4}\times{N}}}$ is reconstructed feature map with U-net structure.

Finally, in the Grapheme-based Predictor, each grapheme is classified with linear classifier $\varphi$ as follows:
\begin{equation}
\textit{$g$} = \textit{$\varphi$}(f_d)\quad{ \in R^{L\times C}}
\end{equation}
where \textit{C} is the number of graphemes in the Hangul.
We use cross entropy loss for the individual grapheme classification.
During the inference, the predicted graphemes are merged into the final output in a form of characters as follows:
\begin{equation}
\textit{$s$} = \textit{h}(\textit{$g^{*}$})
\end{equation}
where $g^{*}\quad{ \in R^{L\times 1}}$ is \textit{top-1} classification output. 
Note that the length of the \textit{s} is $\frac{L}{3}$.
 
There is an exceptional type of characters where the last grapheme doesn't exist. To handle this case, we add a virtual class of \textquotedblleft no-last-consonant\textquotedblright to unify the two different types of characters of [\textit{first consonant, vowel}] and [\textit{first consonant, vowel, last consonant}] structure into a single structure of [\textit{first consonant, vowel, last consonant}].

\section{Experiments}
\subsection{Datasets}
We newly present a benchmark for Hangul OCR which has the intractable number of classes to recognize. 
% 무슨말인지 구체화...
This benchmark reveals the class-imbalance and target-class selecting issues in Hangul OCR. The benchmark dataset is available at \url{https://github.com/mandal4/HangulNet}

\textbf{AI Hub} 
It consists of about 100,000 images of Hangul characters supported by an AI integrated platform operated by the Korea Intelligent Information Society Agency\cite{aihub}. A total of 674,110 text areas are extracted to evaluate the performance of character recognition, excluding character detection. Of these, 10,000 are separated into the test set, and the rest are used as a training data.

%\subsubsection{MLT-h}
\textbf{MLT-h}
MLT dataset \cite{nayef2017icdar2017} was introduced as a part of Robust Reading Competition in ICDAR to resolve the problem of multi-lingual text detection and script identification. We exploit only  the Hangul text regions in the MLT17 test-set for the evaluation, and name it as MLT-h. Note that since we have found many annotation errors in this data set, we rectified those noisy labels.

\textbf{{Standard Foreign Words (SFW)}}
In order to objectively evaluate whether the proposed model effectively alleviates the class-imbalance problem in Korean character recognition, we have synthesized a new dataset containing a large number of minority classes using SynthTiger\cite{yim2021synthtiger}. The dataset contains a total of 18,831 standard foreign words\cite{foreign} that are registered in the National Institute of the Korean Language, which is used only as the test set.

\textbf{{Unseen Characters}}
From the aforementioned SFW dataset, we have selected 72 characters that could not be represented with a common character encoding, and generated an image per character. Then, we measure the performances of various models on this dataset to compare the robustness on the unseen characters.

\subsection{Implementation Details}
We employ 5 transformer layer for the encoder. $C$ is set to 52. Grapheme Merger is simply implemented with UTF-8 unicode.  
We trained all of the benchmark models for 800K iterations with a learning rate of 0.001, and a batch size of 300. The input images are resized into 32 x 128. All experiments are conducted on two Nvidia V100s. 
\subsection{Experimental Settings}
We compare our performance with the Tesseract\cite{smith2007overview}, EasyOCR\cite{easyocr} and more competing methods\cite{cnn_base,yu2020towards}. We reproduce the methods\cite{cnn_base,yu2020towards} and obtain the model with the same training dataset. According to the previous studies, most commonly used 1,549 Hangul characters are chosen as the baseline encoding. In addition, we conduct an ablation to verify the capability of the transformer in modeling the context between consonants and a vowel that defines a character. 

\begin{figure}[htb!]
\begin{center}
\includegraphics[width=1.0\linewidth,height=1.8cm]{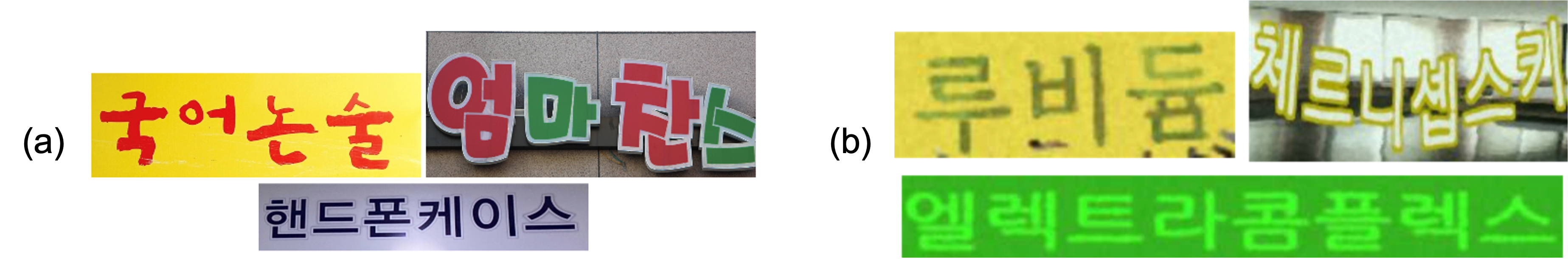}
\end{center}
\caption{Sample images of the datasets: (a) AI Hub covers large-scale of the scene text images in Hangul. (b) SFW contains a number of minority classes and unseen characters from AI Hub. }\label{fig:merged_samples}
\end{figure}

\begin{comment}
\begin{figure}[htb!]
\begin{center}
\includegraphics[width=0.6\linewidth,height=1.5cm]{Figs/aihub_samples.png}
\end{center}
\caption{Examples of the AI Hub dataset. It covers large-scale of the scene text image in Hangul. }\label{fig:aihub_samples}
\end{figure}

\begin{figure}
\begin{center}
\includegraphics[width=0.6\linewidth,height=1.5cm]{Figs/foreign_samples.png}
\end{center}
\caption{Examples of the SFW dataset. It contains a large number of minority classes and even contains unseen classes in AI Hub dataset. }\label{fig:foreign_samples}
\end{figure}
\end{comment}

\subsection{Quantitative Results}
For the quantitative performance comparison, we evaluate the character-level accuracy and the word-level accuracy. Previous studies\cite{} assess the performance only on the selected target-classes, yet,
% describable characters are considered following previous studies. 
we set up an additional experimental setup to simulate real-world scenarios where characters not included in the training set occasionally appear.
For SFW* in Table \ref{Tab:globalResult}, the accuracy is calculated on the whole range of characters covered in SFW dataset. 
% unseen characters that are not included in SFW dataset are considered as incorrect.  s the robustness  experimental setting where especially re-calculate the accuracy where non-describable characters are considered as incorrect to simulate real world scenarios (SFW*)for SFW dataset, . 
The results from Tesseract\cite{smith2007overview} and EasyOCR\cite{easyocr} are not reported as their target-classes are not verifiable. 
% Note that this is not an issue in alphabet recognition.

\begin{table}[]
\begin{center}
\centering
\caption{Experimental comparison of Korean text recognition performance with other methods. The numerical values in the first and second rows of each cell represent the character-level accuracy and the word-level accuracy, respectively. 
% the second of The metrics are in terms of character-level(top) and word-level(bottom) accuracy.
}\label{Tab:globalResult}
\begin{adjustbox}{width=0.45\textwidth}

\begin{tabular}{|c|c|c|c|c|}
\hline
Model                                  & MLT-h & AI Hub & SFW  & SFW* \\ \hline
\multirow{2}{*}{Tesseract\cite{smith2007overview}}             & 13.0   & 9.4      & 12.3    & -    \\
                                       & 11.5   & 8.9      & 7.2    & -    \\ \hline
\multirow{2}{*}{EasyOCR\cite{easyocr}}               & 71.9   & 73.8      & 63.7 & -    \\
                                       & 55.5   & 55.6      & 27.7 & -    \\ \hline
\multirow{2}{*}{Wang et al\cite{cnn_base}}                   & 91.6   & 96.5   & 87.7 & 86.3 \\
                                       & 84.5   & 92.8   & 67.8 & 66.7 \\ \hline
\multirow{2}{*}{SRN\cite{yu2020towards}}           & 93.2   & 97.7   & 89.7 & 88.3 \\
                                       & 89.2   & 95.4   & 72.9 & 71.8 \\ \hline
Ours & 95.8   & 98.1   & 94.0 & 94.0 \\
(w/o Transformer)                                       & 85.9   & 93.0   & 69.2 & 69.2 \\ \hline
\multirow{2}{*}{Ours}                  & \textbf{96.1}   & \textbf{98.8}   & \textbf{96.1} & \textbf{96.1} \\
                                       & \textbf{89.4}   & \textbf{95.5}   & \textbf{77.4} & \textbf{77.4} \\ \hline
\end{tabular}
\end{adjustbox}
\end{center}
\end{table}

As shown in Table \ref{Tab:globalResult}, our approach seamlessly outperforms the existing baselines in all datasets. In particular, the proposed method achieves 4.5\% higher word-level accuracy than the strong baseline\cite{yu2020towards} in SFW where a number of minority classes exist. This shows that our framework is able to successfully classify the graphemes that belong to the minority class. 
% regardless of whether they belong to majority class or minority class. 
For MLT-h and AI Hub, the accuracy gap between our framework and the strong baseline is smaller than that of SFW because the imbalance in these test sets is significant.
Note that the accuracy of minority class is not emphasized when the test-set is imbalanced.

Table \ref{Tab:globalResult} also shows that excluding transformer design in the encoder degrades the word-level accuracy 3.5\%, 2.5\% and 8.2\% in the MLT-h, AI Hub and SFW, respectively. This reveals that transformer captures the context between consonants and a vowel that defines each character. In SFW*, the word accuracy of the strong baseline\cite{yu2020towards} is 1.1\% lower than the SFW counterpart, but the performance of our framework does not get degraded at all. This is because our grapheme encoding enables recognizing characters of all cases in Hangul. 

\begin{table}[h]
\begin{center}
\centering
\caption{Performance comparison on the Unseen Characters dataset. Character-encoding methods\cite{cnn_base,yu2020towards} inevitably fail in classifying any characters that are not included in the training process.}\label{Tab:zeroshotResult2}
\begin{adjustbox}{width=0.45\textwidth}

\begin{tabular}{c|c|c}
\hline
Model               & Encoding unit                   & Accuracy \\ \hline
Wang et al\cite{cnn_base}& character & 0.0 \\ \hline
SRN\cite{yu2020towards}& character & 0.0  \\ \hline
Ours (w/o Transformer)& grapheme & 52.8  \\ \hline
Ours & grapheme & 73.6 \\ \hline
\end{tabular}
\end{adjustbox}
\end{center}
\end{table}

On Unseen Characters dataset, the proposed method shows notable recognition performances. Table \ref{Tab:zeroshotResult2} shows that our framework achieves 73.6\% character-level accuracy. On the other hand, the conventional common character-encoding methods\cite{cnn_base,yu2020towards} can never recognize any unseen characters.

\begin{figure}
\begin{center}
\includegraphics[width=1.0\linewidth,height=4cm]{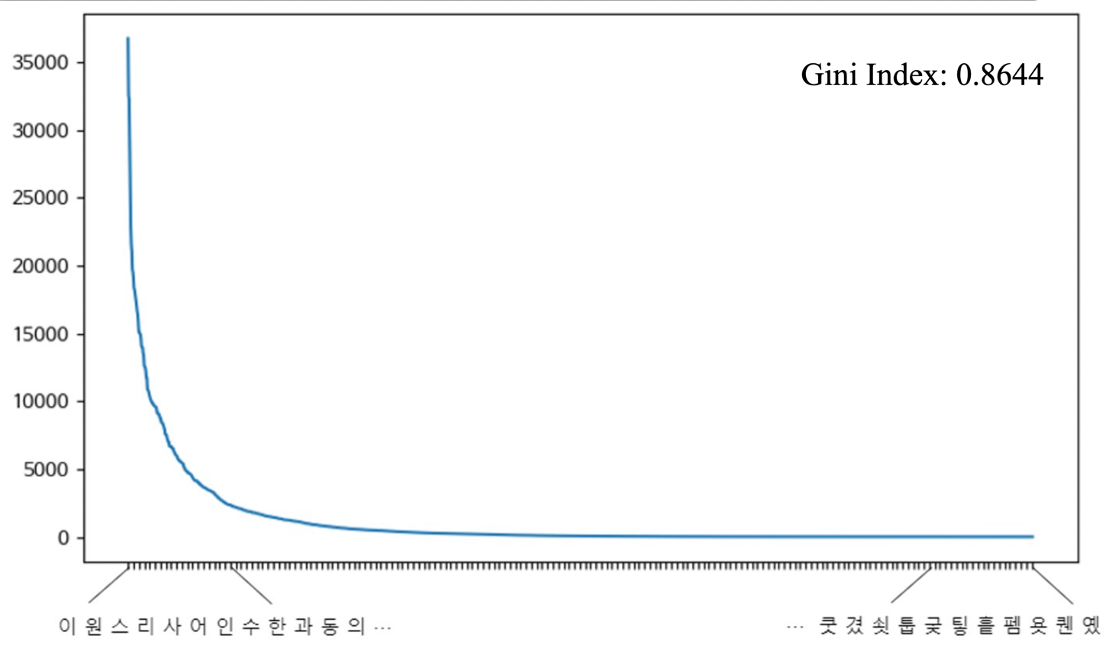}
\end{center}
\caption{Frequency of characters in AI Hub dataset.}\label{fig:sylPDF}
\end{figure}

\begin{figure}
\begin{center}
\includegraphics[width=1.0\linewidth,height=4cm]{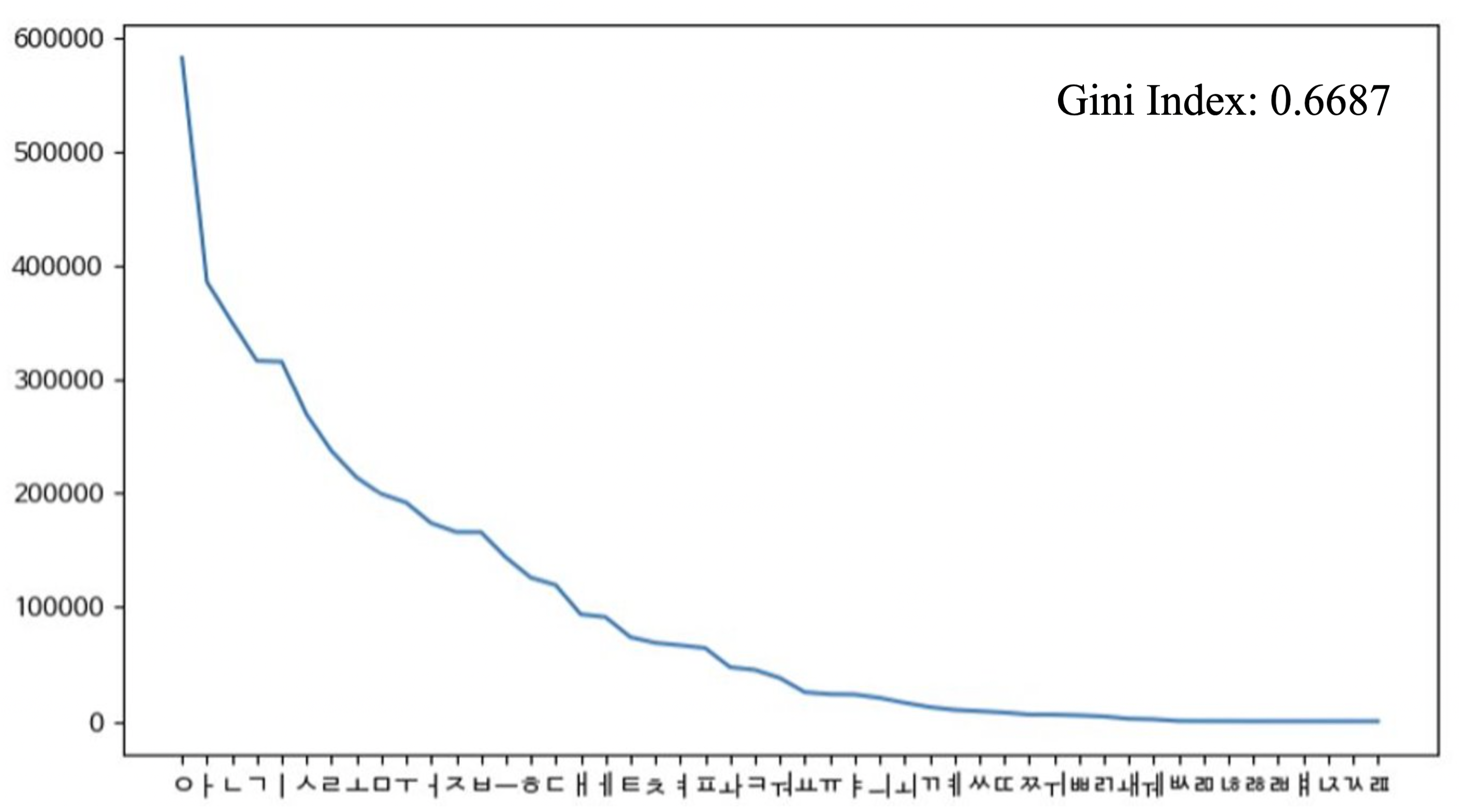}
\end{center}
\caption{Frequency of graphemes in AI Hub dataset. }\label{fig:jamoPDF}
\end{figure}

\begin{figure}[h]
\begin{center}
\includegraphics[width=5cm,height=6cm]{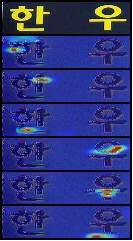}
\end{center}
\caption{Visualization of attention maps to recognize graphemes. The recognition of each grapheme is effectively performed through positional encoding. }\label{fig:attnMap}
\end{figure}

\subsection{Further Analysis}
To verify the effectiveness of our approach on addressing the class-imbalance problem, we plot the frequency of the characters and the graphemes with AI Hub dataset which consists of large-scale real-world Hangul images. As shown in Figure \ref{fig:sylPDF}, there is a huge difference in frequency between the majority character classes and the minority character classes. The imbalance rate of the characters, which is represented with Gini Index, is 0.8644.  On the other hand, when substituted characters with graphemes, the class-imbalance is substantially alleviated and the Gini Index is lowered to 0.6687 (Figure \ref{fig:jamoPDF}).

Additionally, we visualize the attention weight map to validate whether the recognition of each grapheme is effectively performed through positional encoding. Figure \ref{fig:attnMap} shows that the model infers Hangul while following the character construction rule of [\textit{first consonant, vowel, last consonant}]. Note that the last row represents attention map for the class \textquotedblleft no last consonant \textquotedblright.

\section{Conclusion}

In this paper, we propose a grapheme-level Hangul text recognizer with an encoder-decoder structure. With our grapheme-level classification approach, the class-imbalance problem is substantially alleviated by learning to classify the graphemes regardless of whether they are majority or minority characters. Extensive experiments demonstrate the robustness of our approach even in unseen characters that are difficult to handle with conventional methods. Moreover, we newly present a benchmark for Hangul OCR to reveal class-imbalance and target-class selecting issues in Hangul OCR.

% \begin{figure}[t]
%   \centering
%   \fbox{\rule{0pt}{2in} \rule{0.9\linewidth}{0pt}}
%   %\includegraphics[width=0.8\linewidth]{egfigure.eps}

%   \caption{Example of caption.
%   It is set in Roman so that mathematics (always set in Roman: $B \sin A = A \sin B$) may be included without an ugly clash.}
%   \label{fig:onecol}
% \end{figure}

% \begin{figure*}
%   \centering
%   \begin{subfigure}{0.68\linewidth}
%     \fbox{\rule{0pt}{2in} \rule{.9\linewidth}{0pt}}
%     \caption{An example of a subfigure.}
%     \label{fig:short-a}
%   \end{subfigure}
%   \hfill
%   \begin{subfigure}{0.28\linewidth}
%     \fbox{\rule{0pt}{2in} \rule{.9\linewidth}{0pt}}
%     \caption{Another example of a subfigure.}
%     \label{fig:short-b}
%   \end{subfigure}
%   \caption{Example of a short caption, which should be centered.}
%   \label{fig:short}
% \end{figure*}

%------------------------------------------------------------------------

%%%%%%%%% REFERENCES
{\small
\bibliographystyle{ieee_fullname}
\bibliography{egbib}
}

\end{document}